\documentclass[letterpaper]{article} 

\usepackage{aaai25}  
\usepackage{times}  
\usepackage{helvet}  
\usepackage{courier}  
\usepackage[hyphens]{url}  
\usepackage{graphicx} 
\usepackage{natbib}  
\usepackage{caption} 
\usepackage{algorithm}
\usepackage{algorithmic}
\usepackage{newtxmath}
\usepackage{color}

\usepackage{newfloat}
\usepackage{float}
\usepackage{listings}
\usepackage{multirow}
\usepackage{graphicx}
\usepackage{adjustbox}
\usepackage{cite}
\DeclareCaptionStyle{ruled}{labelfont=normalfont,labelsep=colon,strut=off} 
\floatstyle{ruled}
\newfloat{listing}{tb}{lst}{}
\floatname{listing}{Listing}
\title{Riemann-based Multi-scale Attention Reasoning Network for Text-3D Retrieval}
\author {
    Wenrui Li\textsuperscript{\rm 1},
    Wei Han\textsuperscript{\rm 1},
    Yandu Chen\textsuperscript{\rm 1},
    Yeyu Chai\textsuperscript{\rm 1},
    Yidan Lu\textsuperscript{\rm 1},
    Xingtao Wang\textsuperscript{\rm 1}\textsuperscript{\rm 2}\thanks{Corresponding author.},
    Xiaopeng Fan\textsuperscript{\rm 1}\textsuperscript{\rm 2}\textsuperscript{\rm 3}}

\affiliations {
    \textsuperscript{\rm 1}Harbin Institute of Technology \textsuperscript{\rm 2} Harbin Institute of Technology Suzhou Research Institute \textsuperscript{\rm 3}Peng Cheng Laboratory\\
    liwr@stu.hit.edu.cn,2021111641@stu.hit.edu.cn,2021111473@stu.hit.edu.cn,23S136130@stu.hit.edu.cn\\24S103311@stu.hit.edu.cn,xtwang@hit.edu.cn,fxp@hit.edu.cn
}

\usepackage{bibentry}

\begin{document}

\maketitle

\begin{abstract}
Due to the challenges in acquiring paired Text-3D data and the inherent irregularity of 3D data structures, combined representation learning of 3D point clouds and text remains unexplored. In this paper, we propose a novel Riemann-based Multi-scale Attention Reasoning Network (RMARN) for text-3D retrieval. Specifically, the extracted text and point cloud features are refined by their respective Adaptive Feature Refiner (AFR). Furthermore, we introduce the innovative Riemann Local Similarity (RLS) module and the Global Pooling Similarity (GPS) module. However, as 3D point cloud data and text data often possess complex geometric structures in high-dimensional space, the proposed RLS employs a novel Riemann Attention Mechanism to reflect the intrinsic geometric relationships of the data. Without explicitly defining the manifold, RMARN learns the manifold parameters to better represent the distances between text-point cloud samples. To address the challenges of lacking paired text-3D data, we have created the large-scale Text-3D Retrieval dataset T3DR-HIT, which comprises over 3,380 pairs of text and point cloud data. T3DR-HIT contains coarse-grained indoor 3D scenes and fine-grained Chinese artifact scenes, consisting of 1,380 and over 2,000 text-3D pairs, respectively. Experiments on our custom datasets demonstrate the superior performance of the proposed method. Our code and proposed datasets are available at \url{https://github.com/liwrui/RMARN}.

\end{abstract}

\section{Introduction}
Cross-modal retrieval has attracted significant attention due to its effectiveness in aligning multimodal features \cite{wenrui01,wenrui02}. With the recent advancements in applications such as AR/VR and the metaverse, the efficient alignment and processing of point cloud data have become increasingly crucial. Unlike 2D data (such as text) \cite{tang01,tang02,chen01,chen02}, 3D scene data (point clouds) \cite{chu01} provides richer spatial information and is less susceptible to occlusion. Research into Text-3D retrieval, which captures the associations between 3D data and textual descriptions, holds significant potential for the future management and utilization of large-scale 3D data resources. However, the development of this fundamental task faces challenges, mainly due to the difficulty of obtaining paired text-3D data and the inherent irregularity of 3D data structures.
\begin{figure}
	\centering
	\includegraphics[width=1\linewidth]{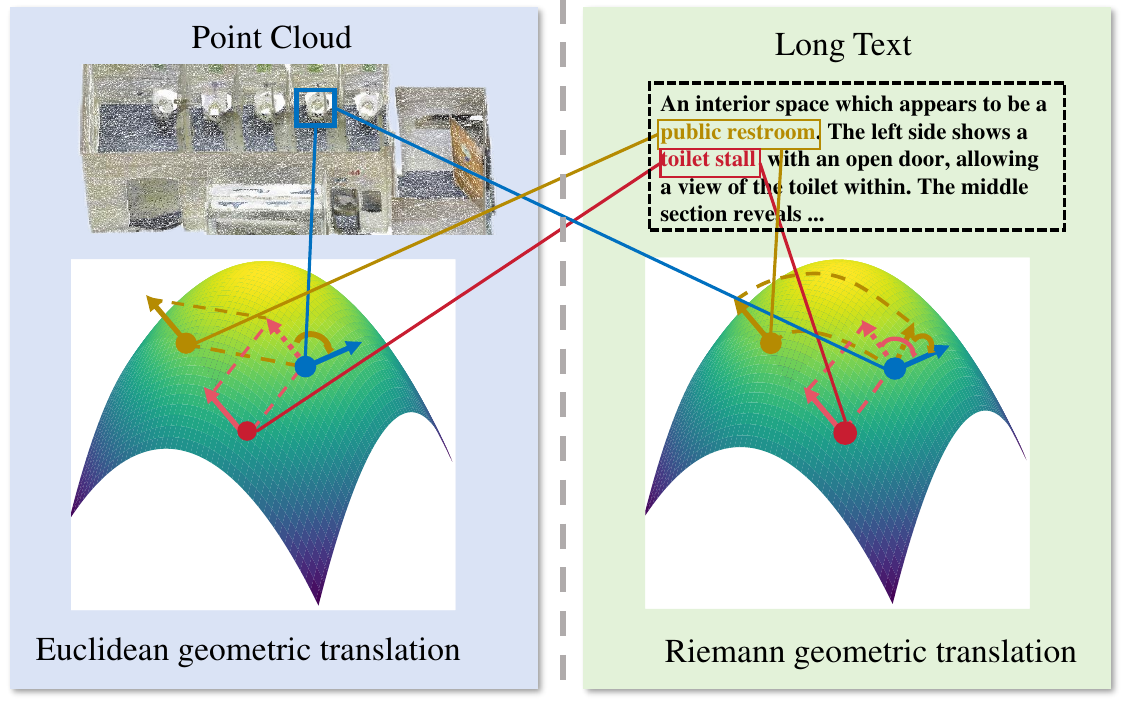}
	\caption{Directly calculating the cosine similarity between two vectors may result in vectors with different meanings at different positions having the same similarity. However, in Riemannian geometry, vector movement conforms to the properties of the manifold, which can mitigate this problem.
}
	\label{fig2}
\end{figure}

Due to the fundamentally different characteristics of text and point cloud data, projecting features into a traditional Euclidean space and comparing them using cosine similarity have significant limitations. This approach fails to adequately capture the complex structures and semantic information inherent in the data, limiting the retrieval model's accuracy and depth of content understanding. To better distinguish the spatial structures of text data and point cloud data, we introduce Riemannian geometry for projecting and aligning cross-modal data features, as illustrated in Fig. 1. Riemannian geometry is particularly effective in handling point cloud data with complex geometric and topological structures, enabling a more accurate representation of the layered nature of point clouds. Since both point cloud and text features are sequences, they can be viewed as fields on specific manifolds. By projecting text and point cloud data onto different manifolds and learning their intrinsic structures on low-dimensional Riemannian manifolds, we can better preserve local information and more accurately represent the complex geometric structures of 3D point cloud data. This approach enhances the model's ability to capture and represent the fine-grained relationships between text and 3D data, resulting in more robust cross-modal retrieval performance.

In this paper, to address the scarcity of paired text-3D data, we developed a large-scale, high-quality open-source dataset named T3DR-HIT, containing over 3,380 pairs of text and point cloud data. The dataset comprises two main parts: one part contains coarse-grained alignments between indoor 3D scenes and text, consisting of 1,380 text-3D pairs; the other part contains fine-grained alignments between Chinese cultural heritage scenes and text, with over 2,000 text-3D pairs. The release of the T3DR-HIT dataset provides robust support for multi-scale text-3D retrieval tasks. Alongside the Riemann Local Similarity (RLS) module, which uses Riemannian geometry to enhance cross-modal data feature alignment, we introduced a Global Pooling Similarity (GPS) module to calculate global similarity between text and point cloud features. To further investigate the low-rank characteristics of text and point cloud data, we proposed a Low-Rank Filter (LRF) module. This module aims to identify sparse correspondences between text and point cloud elements, reducing the model's parameter count while improving its robustness. Furthermore, we effectively integrated the local fine-grained similarity produced by the RLS module, processed with Structured Contextual Pooling (SCP), with the global coarse-grained similarity from the GPS module. This multi-scale similarity calculation strategy not only ensures the model's baseline accuracy but also significantly enhances its ability to recognize and distinguish difficult negative samples. Our main contributions can be summarized as follows:
\begin{itemize}
    \item We used Riemannian geometry to enhance cross-modal feature alignment by projecting features onto low-dimensional Riemannian manifolds, which better capture the complex geometric structures of point cloud data and improve the model's ability to distinguish subtle relationships between text and 3D data.
    \item We proposed a multi-scale similarity calculation strategy that integrates RLS and GPS modules and incorporates an LRF module to reduce model complexity, thereby maintaining baseline accuracy while significantly enhancing the model's ability to recognize challenging negative samples.
    \item We developed and released the T3DR-HIT dataset, comprising 3,380 pairs of text and point cloud data, including coarse-grained alignments of indoor 3D scenes and fine-grained alignments of Chinese cultural heritage scenes.
\end{itemize}

\section{Related Work}
\subsection{Cross-model retrieval}
Cross-modal retrieval has gained significant attention because of the semantic heterogeneity present in multi-modal data \cite{9257382,9169915,LI2022109503,YU2020165,10003241,li2022differentiable,10054421,8954016,9010736,Tang_Sheng_Li_Zhang_Li_Liu_2024,li2024objectsegmentationassistedinterprediction}. A key technique in this field is cross-modal alignment, currently divided into two primary approaches: coarse-grained and fine-grained methods. Coarse-grained methods align data by utilizing a shared embedding space and cosine similarity \cite{9598814,9973278}, as demonstrated by the self-supervised ranking framework \cite{Fu_Li_Mao_Wang_Zhang_2021}. Conversely, fine-grained methods \cite{diao2021similarityreasoningfiltrationimagetext,lee2018stackedcrossattentionimagetext} emphasize cross-modal interactions at the local feature level, such as the semi-supervised learning approach using Graph Convolutional Networks (GCNs) \cite{GCN}. Although effective, the Transformer model is constrained by its large number of parameters. Inspired by the brain’s recurrently connected neurons, RCTRN has shown promising performance \cite{li2023reservoir}. MPARN also addresses annotation uncertainty by modeling visual and textual data as probability distributions \cite{li2024multi}. Cross-modal retrieval has broad applications, such as text-to-video, text-to-audio, and image-text retrieval. However, a significant gap exists in current research on 2D-3D retrieval, despite the substantial structural differences and richer semantic information inherent in 3D data.
\subsection{Indoor scene datasets}
Indoor scene datasets are currently classified into two main categories: scanned scenes and synthetic scenes. The NYU Depth Dataset V2 \cite{SGAC2023}, one of the earliest indoor scene scanning datasets, contains 1,449 RGB-D images across 464 diverse indoor scenes, offering detailed annotations for understanding key surfaces, objects, and support relationships in indoor environments. Other datasets like SUN RGB-D \cite{7298655} and ScanNet \cite{8099744} provide extensive RGB-D data, greatly advancing scene understanding research. Synthetic datasets, compared to scanned scenes, offer the advantage of easy scalability to much larger datasets. SceneNet \cite{7487797}, a framework for indoor scene understanding, generates high-quality annotated 3D scenes by learning from manually annotated real-world datasets like NYUv2. Using a simulated annealing optimization algorithm, SceneNet samples objects from existing 3D object and texture databases, facilitating the creation of an almost unlimited number of new annotated scenes. SUNG \cite{song2016semanticscenecompletionsingle} excels in scalability, while InteriorNet \cite{li2018interiornetmegascalemultisensorphotorealistic} provides highly realistic data. Synthetic datasets, compared to scanned scenes, offer the advantage of easy scalability to much larger datasets. SceneNet \cite{7487797}, a framework for indoor scene understanding, generates high-quality annotated 3D scenes by learning from manually annotated real-world datasets like NYUv2. Using a simulated annealing optimization algorithm, SceneNet samples objects from existing 3D object and texture databases, facilitating the creation of an almost unlimited number of new annotated scenes. SUNG  excels in scalability, while InteriorNet \cite{li2018interiornetmegascalemultisensorphotorealistic} provides highly realistic data. However, none of the aforementioned datasets simultaneously cover both coarse-grained and fine-grained scenes with complex object combinations. To bridge this gap, we propose the T3DR-HIT dataset, a large-scale, high-quality, multi-scale dataset specifically designed for indoor scene understanding.

\begin{figure*}[ht]
	\centering
	\includegraphics[width=1\linewidth]{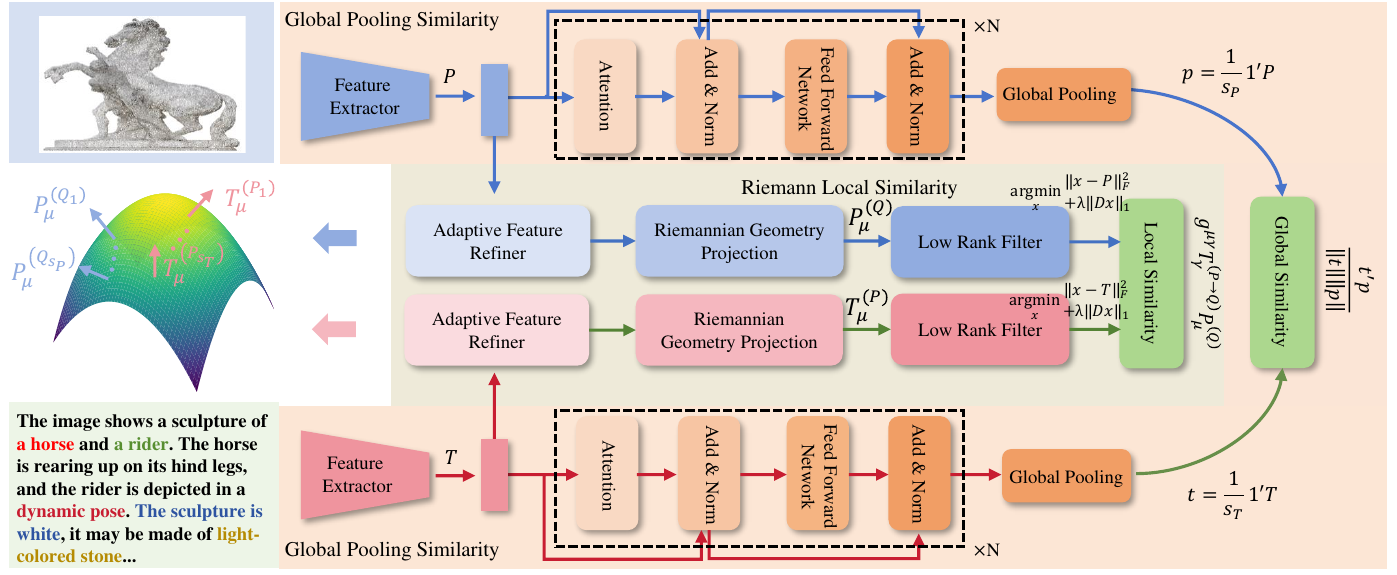}
	\caption{The overall architecture of RMARN proposed in this article. The Global Pooling Similarity module directly calculates the cosine distance between the text feature sequence and the point cloud feature sequence after pooling, while the Riemann Local Similarity module considers the point cloud sequence and the text sequence as two fields on a manifold and calculates the similarity between any two token pairs. Among them, $ T_\mu^{(P_i)} $ and $ P_\mu^{(Q_i)} $ represent the i-th token of the text and point cloud feature sequence, respectively.
}
	\label{fig2}
\end{figure*}

\section{Method}
This section provides a comprehensive description of RMARN, the details is shown in Fig. 2. The core components of RMARN utilize established techniques in natural language processing and 3D data analysis to ensure robust and accurate feature extraction and similarity computation. To extract relevant features from the input data, we use a pre-trained CLIP \cite{CLIP} text encoder optimized for capturing rich textual features across various contexts. This encoder transforms input text into a feature representation, effectively capturing the semantic nuances of the data. Concurrently, we use PointNet \cite{PN}, a well-known method for handling irregular point cloud data, to extract geometric and spatial features from input point clouds. These initial feature extractions are crucial as they form the foundation for subsequent processing. 

After initial feature extraction, features from both modalities undergo further refinement through their respective Adaptive Feature Refiners (AFRs). These refiners are specialized modules designed to enhance the quality of extracted features by adapting them to the specific characteristics of the task at hand. This refinement process results in highly detailed representations, denoted as $ T \in \mathbb{R}^{s_T \times h_T} $ for text and $ P \in \mathbb{R}^{s_P \times h_P}$ for point clouds, where $s_T$ and $s_P$ represent the sequence lengths, and $h_T$ and $h_P$ represent the dimensionality of the features in their respective domains.

We propose a novel similarity computation framework consisting of Riemann Local Similarity (RLS), Global Pooling Similarity (GPS), Similarity Convolution Processor (SCP), and Low-Rank Filter (LRF). We believe this multi-scale similarity computation approach ensures the fundamental accuracy of the model while enhancing its ability to distinguish hard negative pairs. The RLS module processes the refined textual and point cloud features from the AFR, generating a Text-Point Cloud Riemann Attention Map (RAM). Subsequently, the SCP module employs a convolutional network to learn the local similarity of the RAM across $k$ channels, integrating these similarities into the SCP output. Parallel to the RLS, the GPS module conducts global pooling on the AFR outputs and computes cosine similarity, ensuring the model’s performance baseline. Recognizing the low-rank characteristics inherent in both textual and point cloud data, we propose the LRF module as a means of extracting sparse correspondences between text tokens and point cloud tokens. By focusing on these sparse but highly informative correspondences, the LRF module reduces the number of model parameters, thereby improving computational efficiency without sacrificing accuracy.

Finally, the outputs from the SCP and GPS modules are combined to form the final similarity matrix. This matrix integrates the fine-grained local similarities captured by the SCP with the coarse-grained global similarities computed by the GPS, resulting in a comprehensive similarity measure that is both accurate and capable of effectively distinguishing between challenging cross-modal pairs.

\subsection{Adaptive Feature Refiner (AFR) Module}
The textual AFR and point cloud AFR are identical, with each consisting of a stack of six Self-Attention Encoders \cite{c:22}. These AFR modules fine-tune the features of their respective modalities and map them into a common feature space, enabling the subsequent computation of Riemann Attention. Internally, each AFR layer consists of multi-head self-attention (MSA) sub-layers and feed-forward neural network (FFN) sub-layers. Each of these sub-components (MSA and FFN) is encapsulated within residual connections and layer normalization operations. 

The AFR receives text or point cloud inputs, using a scaled dot-product attention mechanism to describe both visual and textual features. The output of the self-attention operator is defined as:
\begin{equation}
    \begin{aligned}
        Attention(\mathbf{Q}, \mathbf{K}, \mathbf{V}) = softmax\left( \frac{\mathbf{Q} \mathbf{K}^T}{\sqrt{d_e}} \right) \mathbf{V},\\ Att(\mathbf{X}) = Attention(\mathbf{X} \mathbf{W}_Q, \mathbf{X} \mathbf{W}_K, \mathbf{X} \mathbf{W}_V),
    \end{aligned}
\end{equation}
where \( \mathbf{W}_Q \in \mathbb{R}^{d \times d_e} \), \( \mathbf{W}_K \in \mathbb{R}^{d \times d_e} \), and \( \mathbf{W}_V \in \mathbb{R}^{d \times d_e} \) represent the learnable linear transformations for the query, key, and value, respectively. \( d_e \) indicates the dimensionality of the embedding space. We utilize a compact feed-forward network (FFN) to extract features, which are already integrated into more extensive representations. The FFN is composed of two nonlinear layers:

\begin{equation}
    \begin{aligned}       
        FFN(\mathbf{X}_i) = \text{GELU}(\mathbf{X}_i \mathbf{W}_1 + \mathbf{b}_1) \mathbf{W}_2 + \mathbf{b}_2,\\
        GELU(x) = \epsilon x \left(1 + \tanh \left[ \sqrt{\frac{2}{\pi}} (x + \rho x^3) \right] \right),
    \end{aligned}
\end{equation}
where \( \epsilon \) and \( \rho \) are hyperparameters, \( \mathbf{X}_i \) represents the \( i \)-th vector of the input set, \( \mathbf{W}_1 \) and \( \mathbf{W}_2 \) are learnable weight matrices, and \( \mathbf{b}_1 \) and \( \mathbf{b}_2 \) are bias terms. The GELU activation function can enhance the model's generalization capabilities.A complete encoding layer \( A_i \) (\( i = 1, \ldots, n \)) can be described as follows:
\begin{equation}
    \begin{aligned}
    \mathbf{S} = Add \& Norm(Att(\mathbf{X})),  \\
    \hat{\mathbf{X}} = Add \& Norm(FFN(\mathbf{S})),
    \end{aligned}
\end{equation}
where \text{Add \& Norm} includes a residual connection and layer normalization. The multi-layer encoder \( A_i \) (\( i = 1, \ldots, n \)) is constructed by stacking these encoding layers sequentially, with the input of each layer being derived from the output of the preceding layer. In the AFR, stacking multiple encoder layers enables the automatic adjustment of weights between features, ensuring that crucial ones receive greater attention. This adaptive feature enhancement makes the model more flexible and efficient in handling complex, high-dimensional text and point cloud data, thereby improving the accuracy of subsequent similarity computations.

\subsection{Riemann Local Similarity (RLS) Module}
The text feature $ \mathbf{T} \in \mathbb{R}^{s_T \times h_T} $ and the point cloud feature $ \mathbf{P} \in \mathbb{R}^{s_P \times h_P} $ are both represented as vector sequences, allowing them to be considered as samples originating from two distinct fields distributed across a manifold at specific loci. To facilitate subsequent derivations, we will adopt the Einstein summation convention, using $ \mathbf{T}_{\mu} $ and $ \mathbf{P}_{\mu} $ as predefined notational constructs to denote the characteristics of text and point clouds pertaining to a specific token.

In Riemannian geometry, directly assessing the similarity between two tensors at different positions is not practically meaningful. To quantify the similarity between tensors at distinct points within two fields, it is essential to first transport them to the same location. This process involves transporting the tensor from the text field at point $ P $ to point $ Q $, accomplishing this by leveraging the connection $ \Gamma $ alongside the displacement $ dx $:
\begin{equation} \label{trans}
    \begin{aligned}
        \mathbf{T}_{\mu}^{(P \rightarrow Q)} &= \mathbf{T}_{\mu}^{(P)} + \Gamma_{\mu\gamma}^{\alpha} \mathbf{T}_{\alpha}^{(P)} {dx}^{\gamma} \\ 
        &= \mathbf{T}_{\alpha}^{(P)}(\delta_\mu^\alpha  + \Gamma_{\mu\gamma}^{\alpha} {dx}^{\gamma}),
    \end{aligned}
\end{equation}
where $ \delta_\mu^\alpha $ is the Kronecker symbol. Calculate the similarity between two tensors at the same position after translation using dot product:

\begin{equation} \label{sim}
    sim(\mathbf{T}_{\mu}^{(P \rightarrow Q)}, \mathbf{P}_\mu^{(Q)} ) = g^{\mu \gamma} \mathbf{T}_{\mu}^{(P \rightarrow Q)} \mathbf{P}_\gamma ^{(Q)},
\end{equation}
where $ g^{\mu \gamma} $ is the metric of the manifold.

By substituting equation \ref{trans} into equation \ref{sim}, we can obtain:

\begin{equation}
    \begin{aligned}
        sim(\mathbf{T}_{\mu}^{(P \rightarrow Q)}, \mathbf{P}_\mu ^{(Q)} ) &= g^{\mu\gamma}(\delta^{\alpha}_{\mu} + \Gamma_{\mu\gamma}^{\alpha} {dx}^{\gamma})\mathbf{T}_{\alpha}^{(P)} \mathbf{P}_\gamma ^{(Q)}
        \\ &= \Omega^{\alpha\gamma}\mathbf{T}_{\alpha}^{(P)} \mathbf{P}_\gamma ^{(Q)},
    \end{aligned}
\end{equation}
where $ \Omega^{\alpha\gamma} = g^{\mu\gamma}(\delta^{\alpha}_{\mu} + \Gamma_{\mu\gamma}^{\alpha} {dx}^{\gamma}) $ is a two-dimensional tensor, and depends on the position of points $P$ and $Q$.

To simplify the parameters, the $ \mathbf{T}_\alpha\Omega^{\alpha\gamma}\mathbf{P}_\gamma $ term related to PQ position is approximately decomposed into a PQ position independent term $ t'\mathbf{Q} p $ and a position only term $ e $, and $\mathbf{Q}$ is matrix decomposed to obtain:
\begin{equation}
    \begin{aligned}
        &\mathbf{T}_\alpha\Omega^{\alpha\gamma}\mathbf{P}_\gamma = t'\mathbf{Q} p + e^{(P,Q)}
        \\ &=t'A'Bp + e^{(P,Q)}=(At)'(Bp) + e^{(P,Q)},
    \end{aligned}
\end{equation}
where $ t $ and $ p $ represent specific token features within the text feature sequence $ \mathbf{T} $ and the point cloud feature sequence $ \mathbf{P} $, respectively. Using the aforementioned equation, we can compute the local (token-level) similarity between any pair of text and point cloud tokens. This local similarity metric can then be used to enhance attention to intricate details when calculating the overall global similarity.

\subsection{Similarity Convolution (SC) Module}
On any manifold, the local similarity between any pair of text and point cloud tokens can be represented as a local similarity matrix. To enhance the similarity measurement from multiple perspectives, we can compute local similarity matrices on multiple ($k$) manifolds, resulting in a similarity map $ \mathbf{M} \in \mathbb{R}^{k \times s_1 \times  s_2} $ with $k$ channels. We can use convolutional network $ \mathbf{C} $ to construct a mapping that converts the $k$ local similarity matrices to the total similarity $ s \in \mathbb{R} $:
\begin{equation}
    s = \mathbf{C} \circ \mathbf{M}.
\end{equation}

\subsection{Low Rank Filter (LRF) Module}
Given the inherent constraints of compressing data within the model, redundant information inevitably persists within both point cloud feature sequences and text feature sequences, hindering the model's generalization capabilities and exacerbating computational intricacies. Consequently, it becomes imperative to leverage low-rank priors \cite{hu2021lora} as a means of eliminating this redundant information.

When given the original feature map $\mathbf{M}$ containing redundant information, we can use the following equation to extract the low rank component $\mathbf{X}$ from it:

\begin{equation}
    \mathbf{X}=arg\min_{x} \left \{ {\left \| \mathbf{M}-x \right \|^{2}_{F} + \lambda \left \| \mathbf{D}x \right \|_{1} } \right \}, 
\end{equation}
where $\lambda $ is the regularization coefficient that balances sparse loss and data restoration loss. Assuming $ \mathbf{D} $ is orthogonal, then the minimization problem has a closed solution $\mathbf{X}=\mathbf{D}^ {-1}soft (\mathbf{DM}, \ lambda) $, where $soft$ is the soft interval function:

\begin{equation}
    soft(x,\lambda )=\left\{\begin{matrix}
     x-\lambda, & x>\lambda;\\
     x+\lambda, & x<-\lambda;\\
     0, & otherwise.
    \end{matrix}\right.
\end{equation}

This article uses neural networks to approximate the mapping of $\mathbf{D} $. Since the total similarity $ s $ is a function of $X$, it is:
\begin{equation}
    s = \mathbf{CX} = \mathbf{CD}^{-1}soft(\mathbf{DM}, \lambda ).
\end{equation}

Therefore, a complete neural network can be used to simultaneously approximate $\mathbf{CD} ^ {-1} $without explicitly approximating $\mathbf{C} $ and $\mathbf{D} ^ {-1} $separately.

\subsection{Comparative Learning Loss for RMARN}
RMARN aims to maximize the similarity between paired point clouds and text samples, while minimizing the similarity between unmatched samples. Consequently, it adopts a similar contrastive loss framework as CLIP, leveraging softmax to derive the retrieval pairing probability, which is grounded on the computed similarity between point cloud text pairs $ (p,t) $ sampled from our dataset $ D $. The specific formulation is outlined below:
\begin{equation}
    \mathcal{L} = - E_{p(t,p|D)}\left [ \alpha_1 log(\frac{e^{\frac{f(t,p)}{\tau_1}}}{ {\textstyle \sum_{p'} {e^{\frac{f(t,p')}{\tau_1}}}}}) + \alpha_2 log(\frac{e^{\frac{f(t,p)}{\tau_2}}}{ {\textstyle \sum_{t'} {e^{\frac{f(t',p)}{\tau_2}}}}}) \right ],   
\end{equation}
where $f$ is a similarity calculation function implemented by RMARN, $\alpha_1$ and $\alpha_2$ are served as hyperparameters, used to balance the directional focus when retrieving text and point clouds from each other.

\begin{figure*}[ht]
	\centering
	\includegraphics[width=1\linewidth]{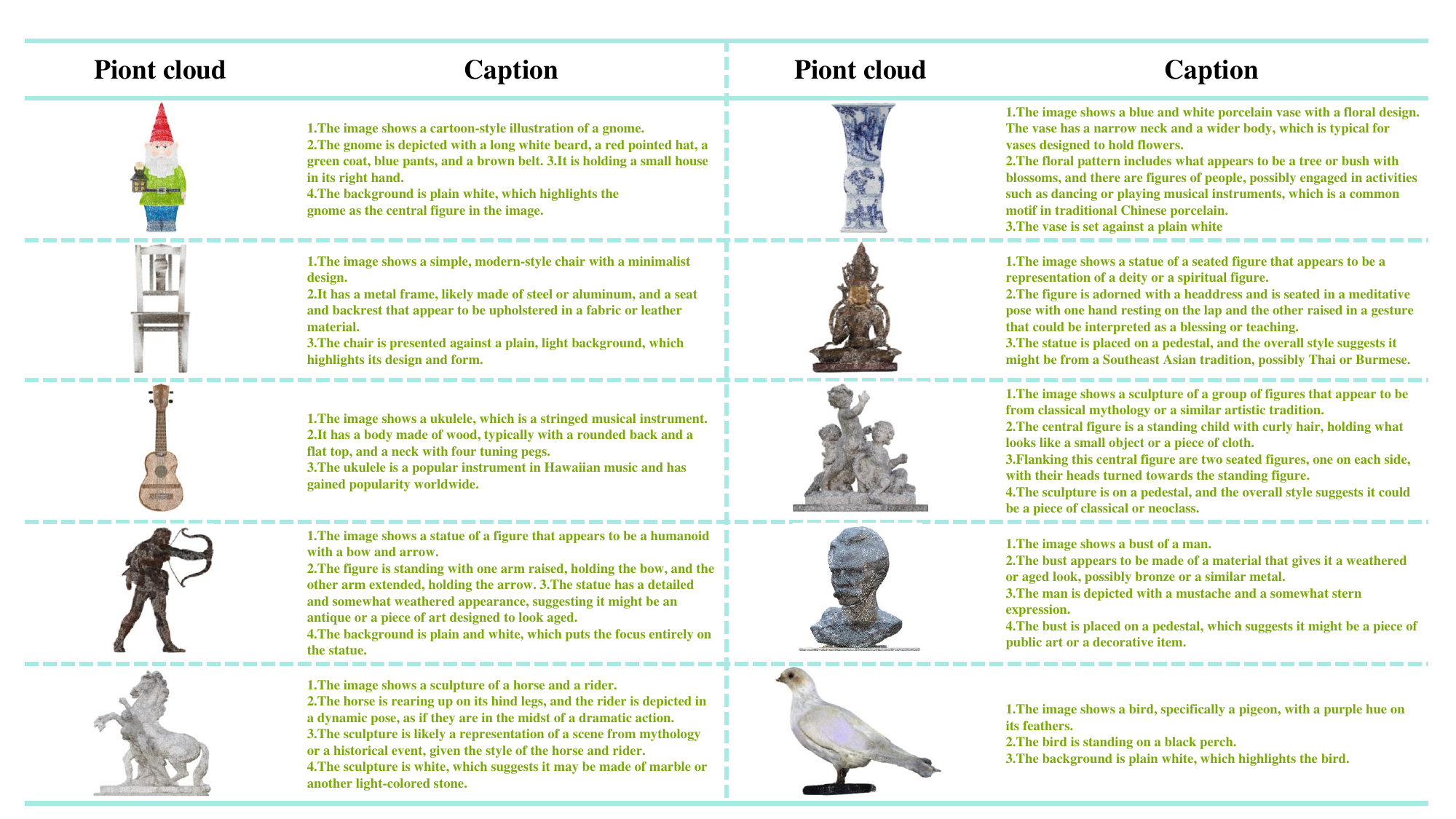}
	\caption{
 Examples of text point cloud pairs in The Elephant Meta Dataset. Each point cloud describes a fine-grained 3D object, and each point cloud corresponds to a caption consisting of 3 or more sentences that describe the specific content of the point cloud in natural language.
}
	\label{fig2}
\end{figure*}

\section{Experiments}
We conducted comparative experiments on the T3DR-HIT dataset, utilizing different text and point cloud feature extractors while keeping the retrieval framework unchanged. The experimental results demonstrated the superior retrieval performance of our model.

\subsection{Datasets and Evaluation Metric}
T3DR-HIT is a comprehensive, large-scale Text-3D Retrieval dataset designed to facilitate research and development in cross-modal retrieval, particularly involving textual descriptions and 3D spatial data. The dataset comprises over 3,380 pairs of text and point cloud data, making it one of the most extensive resources available for studying the interaction between these two modalities. T3DR-HIT is divided into two distinct segments to accommodate different levels of granularity in 3D scene representation: coarse-grained Indoor 3D Scenes and fine-grained Chinese Artifact Scenes.

\paragraph{Indoor 3D scenes} 
Building on the open-source Stanford 2D-3D-Semantics Dataset, we developed the Indoor Text-Point Pairs dataset—a novel resource aimed at enhancing cross-modal research in indoor scene understanding. We utilized GPT-4o, an advanced language model, to generate descriptive captions for panoramic images in the Stanford 2D-3D-Semantics dataset. Each room was annotated with multiple captions, which were then carefully paired with corresponding point clouds, creating a robust dataset linking textual descriptions with 3D spatial data. The Stanford 2D-3D-Semantics Dataset is a comprehensive collection focused on large-scale indoor environments. It offers a wide array of co-registered modalities across the 2D and 3D domains, making it an invaluable resource for tasks in computer vision and machine learning. The dataset spans over 6,000 square meters of indoor space and includes over 70,000 high-resolution RGB images. Alongside these images, the dataset provides corresponding depth maps, surface normals, and semantic labels, which are critical for understanding the geometric structures of indoor scenes. 

Additionally, the dataset includes global XYZ images available in both standard formats and 360-degree equirectangular projections, offering a comprehensive view of spatial relationships within the scenes. Camera metadata is also included, ensuring accurate alignment and registration across different modalities. The 3D component of the dataset is equally rich, featuring both raw and semantically annotated 3D meshes and point clouds. These 3D representations are crucial for tasks such as object recognition, scene segmentation, and spatial reasoning in indoor environments.

\paragraph{Chinese artifact scenes}
The Elephant Meta Dataset, curated by Henan Broadcasting and Television Station, is a specialized collection featuring detailed mesh data of Chinese ancient artifacts, including significant items like the blue and white porcelain emblem, Tang Dynasty court lady figurine, coiled dragon stone pillar, and stone-seated qilin. As shown in Fig. 3, this dataset provides a unique opportunity to explore cultural heritage through advanced 3D data analysis and cross-modal learning. To utilize this dataset for research and development, we employed Open3D, a versatile library for 3D data processing, to visualize the colored mesh data. By rendering these meshes in Open3D, we captured high-quality 2D screenshots, converting the 3D artifacts into 2D visual representations. These images serve as the foundation for further analysis and caption generation.

To generate descriptive captions for these 2D images, we utilized the LLaVA large language model, specifically the llava-v1.6-mistral-7b-hf version. This model excels at interpreting and describing visual content, enabling us to produce accurate and contextually relevant captions for each artifact image. These captions encapsulate the visual and cultural details in the images, bridging the gap between visual data and textual descriptions. In addition to the 2D images and captions, we generated point cloud data by uniformly sampling 100,000 points from the surface of each colored mesh. This process captures the geometric structure and surface details of the artifacts, translating rich 3D information into a format suitable for further computational analysis. This process captures the geometric structure and surface details of the artifacts, translating rich 3D information into a format suitable for further computational analysis. This paired dataset serves as a valuable resource for tasks involving cross-modal learning, cultural heritage preservation, and the study of Chinese ancient artifacts.

\begin{table*}[h!]
\centering
\setlength{\tabcolsep}{4pt} 
\renewcommand{\arraystretch}{1.2} 
\begin{tabular}{cccccccccccc}
\hline \hline
\multirow{2}{*}{Model Type} & \multicolumn{2}{c}{Feture Extracter} & \multicolumn{4}{c}{T3DR-HIT Dataset}                   & \multicolumn{5}{c}{Hyper-parameters}                                 \\ \cline{2-12} 
                            & Text           & PC                  & R@1         & R@5         & R@10        & Rsum         & Low Rank     & Epochs       & Batch Size  & Nhead       & SA Layer   \\ \hline
\multirow{4}{*}{Fast Model} & GNN            & PointNet            & 0           & 1           & 3           & 4            & 128          & 20           & 32          & 16          & 4          \\
                            & CLIP           & PointNet            & 0           & 0           & 5           & 5            & 128          & 20           & 32          & 16          & 4          \\
                            & GNN            & PointNet++          & 1           & 4           & 6           & 11           & 128          & 20           & 32          & 16          & 4          \\
                            & CLIP           & PointNet            & 0           & 8           & 8           & 16           & 128          & 20           & 32          & 16          & 4          \\ \hline
\multirow{5}{*}{Base Model} & CLIP           & PointNet            & 21          & 31          & 37          & 89           & 256          & 100          & 32          & 16          & 6          \\
                            & CLIP           & PointNet            & 13          & 39          & 47          & 99           & 256          & 100          & 64          & 16          & 6          \\
                            & CLIP           & PointNet            & 19          & 50          & 53          & 122          & 256          & 80           & 32          & 32          & 6          \\
                            & BERT           & PointNet++          & 25          & 58          & 62          & 145          & 256          & 100          & 64          & 32          & 6          \\
                            & \textbf{BERT}  & \textbf{PointNet++} & \textbf{31} & \textbf{61} & \textbf{69} & \textbf{161} & \textbf{256} & \textbf{100} & \textbf{64} & \textbf{32} & \textbf{8} \\ \hline \hline
\end{tabular}
\caption{Comparison of different RMARN configurations on T3DR-HIT dataset.}
\end{table*}

\begin{table}[h!]
\centering
\setlength{\tabcolsep}{6pt} 
\renewcommand{\arraystretch}{1.2} 
\begin{tabular}{ccccc}
\hline
\multirow{2}{*}{Model} & \multicolumn{4}{c}{T3DR-HIT Dataset}                   \\ \cline{2-5} 
                       & R@1         & R@5         & R@10        & Rsum         \\ \hline
W/o GPS                & 16          & 24          & 35          & 75           \\
W/o AFR and RLS        & 23          & 56          & 61          & 140          \\
W/o RLS                & 28          & 55          & 65          & 148          \\
W/o AFR                & 28          & 58          & 67          & 153          \\ \hline
\textbf{RMARN (ours)}  & \textbf{31} & \textbf{61} & \textbf{69} & \textbf{161} \\ \hline
\end{tabular}
\caption{Ablation study on T3DR-HIT dataset.}
\end{table}

\subsection{Implementation Details}
We selected the CLIP text encoder and PointNet as the baseline feature extractors for handling the text and point cloud modalities, respectively. These choices are based on the proven capabilities of CLIP in capturing rich semantic information from textual data and PointNet's effectiveness in processing irregular 3D point clouds. The encoder employs \textit{LayerNorm} for normalization, ensuring stable and consistent scaling of the input data. For the activation function, we utilized \textit{GELU} (Gaussian Error Linear Unit) with the parameters $\epsilon=0.5$ and $\rho=0.044715$ which balances non-linearity and smooth gradient flow. A dropout rate of 0.1 is applied to prevent overfitting by randomly zeroing out a fraction of the neurons during training. Both the \textit{Attention} layer and the Feed-Forward Network (\textit{FFN}) in the self-attention encoder are configured with a dimensionality of 512. This dimensional setting ensures that the model can capture complex relationships within the data without excessively increasing computational costs. We trained the model for 100 epochs, utilizing the \textit{Adam} optimizer, which is well-regarded for its ability to adapt learning rates during training. The learning rate was set to 0.008, providing a balance between making steady progress and avoiding potential overshooting of minima. The $\beta$ for the \textit{Adam} optimizer were configured as (0.91, 0.9993).

\subsection{Comparison Experiment}
The Table 1 summarizes the performance of various models on the T3DR-HIT dataset, including their respective hyperparameter configurations. In the Fast Model category, the GNN-based models (PointNet and PointNet++) show poor retrieval performance across all metrics (R@1, R@5, R@10), especially with near-zero R@1 scores, highlighting weak retrieval accuracy for the top candidate. The CLIP-based models perform slightly better, but their overall performance remains suboptimal. 

In contrast, the Base Model category shows that BERT-based models, especially the BERT + PointNet++ combination, significantly outperform others, achieving an R@10 score of 161, which is much higher than that of other models. Although the CLIP-based models show improvements over the Fast Models, they still lag considerably behind the BERT-based models. The superior performance of the BERT + PointNet++ model can be attributed to its more sophisticated hyperparameter settings, including a higher Low Rank value (256), larger batch size (64), increased number of attention heads (32), and additional SA layers (8), which enhance its feature extraction capabilities and overall retrieval accuracy. This suggests that the BERT + PointNet++ model has substantial potential for 3D object retrieval tasks. The hyperparameter configurations of the models highlight key differences that contribute to their varying performances. These configurations suggest that more complex and well-tuned hyperparameters are crucial for improving the effectiveness of 3D object retrieval tasks.


\subsection{Ablation study}
\subsubsection{Effectiveness of different model components.} The results in Table 2 show a significant drop in performance when the GPS module is removed, with Rsum decreasing from 161 to 75, indicating the crucial role of the GPS module in the model's overall performance. When both the AFR and RLS modules are removed, the model's performance partially recovers to an Rsum of 140, which is still significantly lower than the full model's Rsum of 161. When only the RLS module is removed, the model's R@1 performance remains unchanged, but the accuracies at R@5 and R@10 decrease, leading to an Rsum of 148. Removing the AFR module is smaller, but it still causes Rsum to drop to 153.
\begin{figure}[ht]
	\centering
	\includegraphics[width=0.45\linewidth]{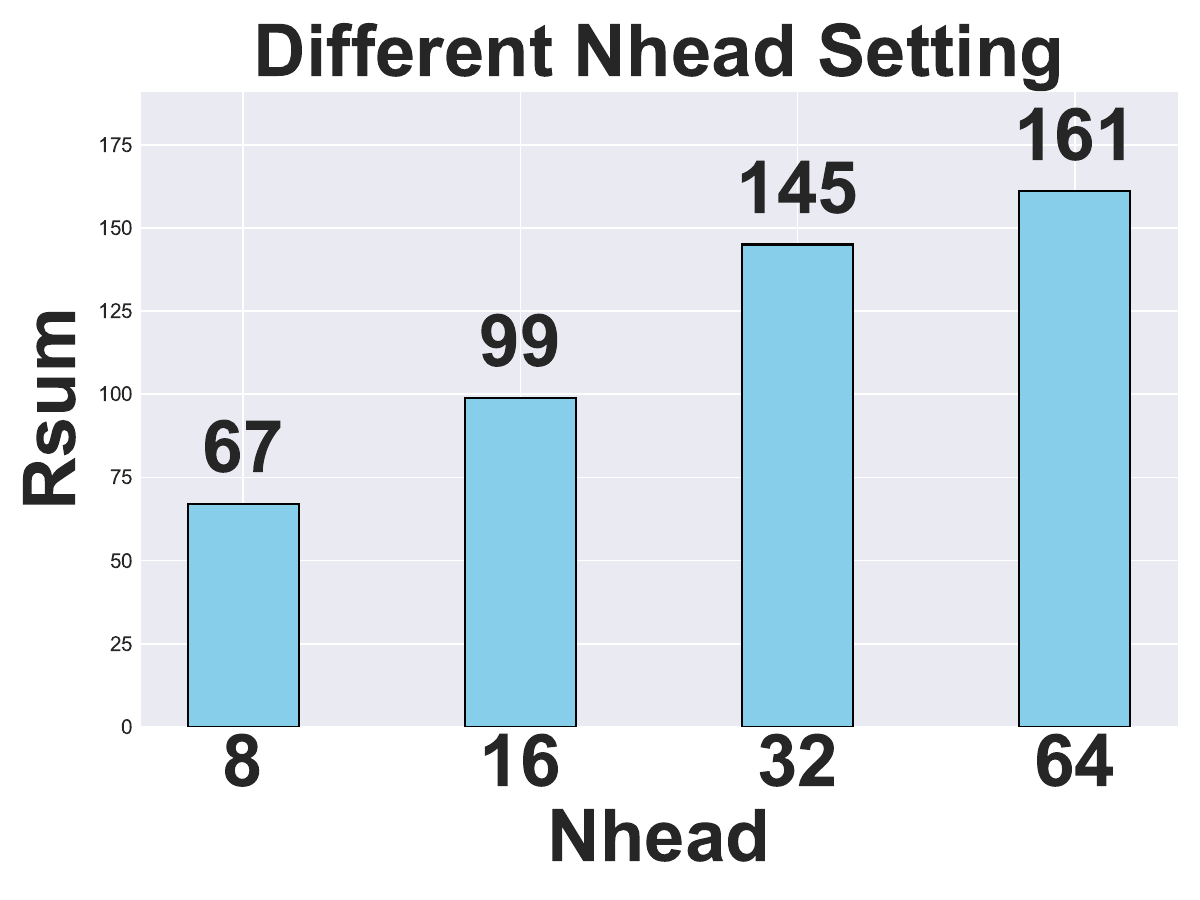}
 \includegraphics[width=0.45\linewidth]{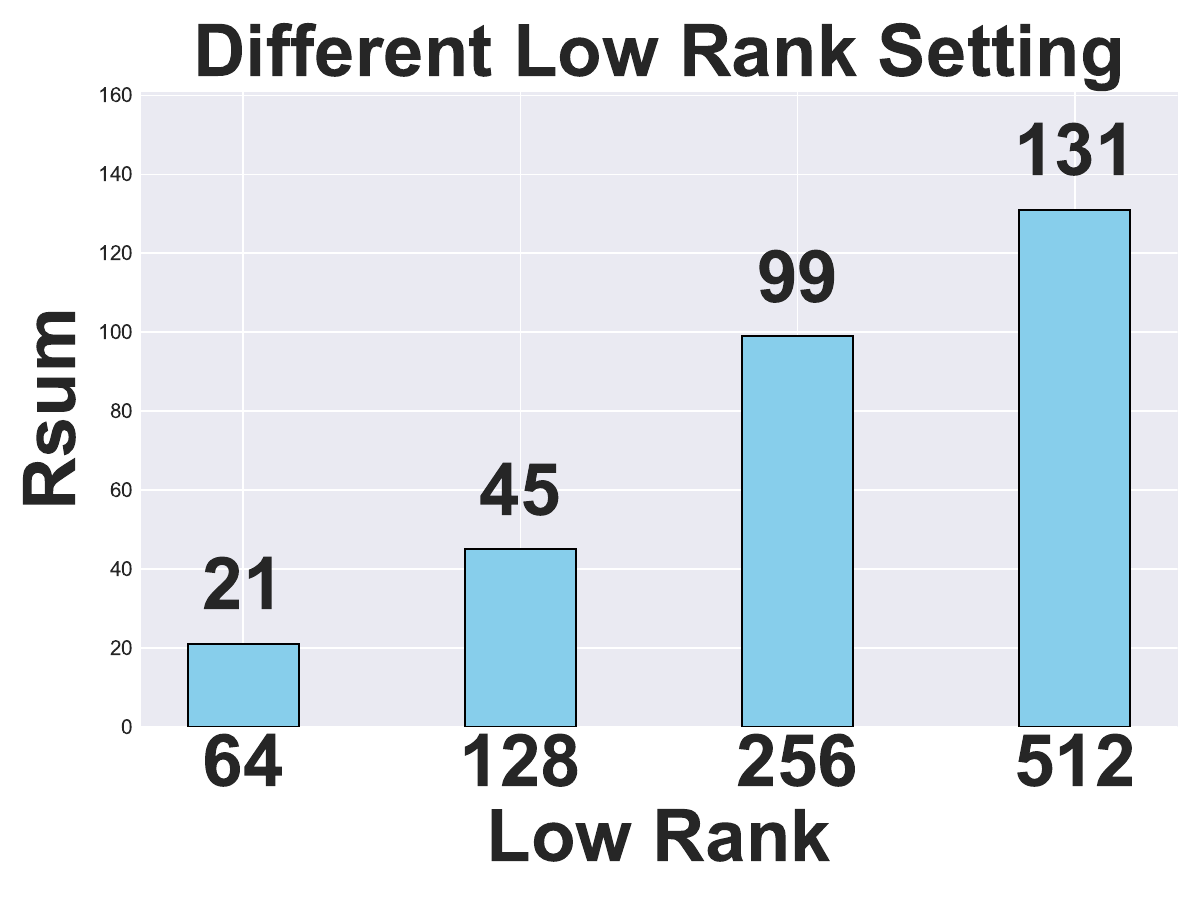}
	\caption{ Impact of different Nhead and low rank settings.
}
	\label{fig2}
\end{figure}
\subsubsection{Effectiveness of different Nhead and low rank setttings.} Fig. 4 illustrates the impact of varying Nhead and low-rank settings on model performance. In the left graph, Rsum increases from 67 to 161 as Nhead rises from 8 to 64, indicating that a greater number of attention heads enhances the overall performance. The right graph shows Rsum improving from 21 to 131 as the low-rank value increases from 64 to 512, suggesting that higher low-rank settings enhance the model's representational capacity. Therefore, the ablation study demonstrates that increasing both Nhead and low-rank parameters results in improved model performance.
\section{Conclusion}
In this work, we introduce RMARN, a novel Riemann-based Multi-scale Attention Reasoning Network, specifically designed for the challenging task of text-3D retrieval. By leveraging the Adaptive Feature Refiner (AFR) to enhance representations of both text and point cloud data, and incorporating the Riemann Local Similarity (RLS) and Global Pooling Similarity (GPS) modules, our approach effectively captures the complex geometric relationships inherent in high-dimensional spaces. The proposed Riemann Attention Mechanism enables RMARN to learn manifold parameters that accurately reflect the intrinsic distances between text-point cloud samples without requiring explicit manifold definitions. To address the scarcity of paired text-3D data, we developed the T3DR-HIT dataset, the first large-scale dataset of its kind, comprising 3,380 diverse text-point cloud pairs across both coarse-grained indoor scenes and fine-grained Chinese artifacts. Extensive experiments conducted on our custom datasets validate RMARN's efficacy, demonstrating its superior performance over existing methods. We believe our contributions pave the way for further advancements in cross-modal retrieval tasks, particularly in the underexplored domain of text-3D retrieval.
\section{Acknowledgments}
This work was supported in part by the National Key R\&D Program of China (2021YFF0900500), the National Natural Science Foundation of China (NSFC) under grants 62441202, U22B2035, 20240222, and the Fundamental Research Funds for the Central Universities under grants HIT.DZJJ.2024025.

\end{document}